\documentclass[letterpaper]{article} 
\usepackage{aaai24}  
\usepackage{times}  
\usepackage{helvet}  
\usepackage{courier}  
\usepackage[hyphens]{url}  
\usepackage{graphicx} 
\usepackage{natbib}  
\usepackage{caption} 
\usepackage{algorithm}
\usepackage{algorithmic}

\usepackage{newfloat}
\usepackage{listings}
\DeclareCaptionStyle{ruled}{labelfont=normalfont,labelsep=colon,strut=off} 
\floatstyle{ruled}
\newfloat{listing}{tb}{lst}{}
\floatname{listing}{Listing}
\usepackage{amsthm}
\newtheorem{theorem}{Theorem}
\usepackage{amsmath}
\usepackage{amssymb}
\usepackage{adjustbox}
\usepackage{booktabs}
\usepackage{makecell}
\usepackage{multirow}
\usepackage{pifont}
\newcommand{\cmark}{\ding{51}}
\newcommand{\xmark}{\ding{55}}

\title{Density Matters: Improved Core-set for Active Domain Adaptive Segmentation}
\author {
    Shizhan Liu\textsuperscript{\rm 1}\equalcontrib,
    Zhengkai Jiang\textsuperscript{\rm 2}\equalcontrib,
    Yuxi Li\textsuperscript{\rm 2}\equalcontrib,
    Jinlong Peng\textsuperscript{\rm 2},
    Yabiao Wang\textsuperscript{\rm 2\dag},
    Weiyao Lin\textsuperscript{\rm 1}\footnote{Corresponding author.}
}
\affiliations {
    \textsuperscript{\rm 1}Shanghai Jiao Tong University\\
    \textsuperscript{\rm 2}Tencent Youtu Lab\\
    shanluzuode@sjtu.edu.cn, zhengkjiang@tencent.com, cookieyxli@gmail.com,
    jeromepeng@tencent.com,
    caseywang@tencent.com,
    wylin@sjtu.edu.cn
}

\begin{document}

\maketitle

\begin{abstract}
Active domain adaptation has emerged as a solution to balance the expensive annotation cost and the performance of trained models in semantic segmentation. However, existing works usually ignore the correlation between selected samples and its local context in feature space, which leads to inferior usage of annotation budgets. In this work, we revisit the theoretical bound of the classical Core-set method and identify that the performance is closely related to the local sample distribution around selected samples. To estimate the density of local samples efficiently, we introduce a local proxy estimator with Dynamic Masked Convolution and develop a Density-aware Greedy algorithm to optimize the bound. Extensive experiments demonstrate the superiority of our approach. Moreover, with very few labels, our scheme achieves comparable performance to the fully supervised counterpart.
\end{abstract}

\section{Introduction}
Semantic segmentation has become increasingly crucial in various applications, such as autonomous driving~\cite{teichmann2018multinet}, virtual try-on~\cite{virtual_try_on}, and smart healthcare~\cite{shi2020review}. In recent years, remarkable progress has been made in this field~\cite{chen2017deeplab, he2017mask, chen2018encoder, maskformer}. However, annotating each pixel in an image is extremely costly. Thus, domain adaptation techniques have been introduced to overcome the high cost of annotation~\cite{chang2019all, cheng2021dual, zhang2021prototypical, liu2021bapa}. Unfortunately, the performance of unsupervised domain adaptation methods still lags far behind that of fully supervised training on the target domain. Therefore, active domain adaptation has emerged as a solution to balance the expensive annotation cost and the performance of the trained segmentation model, which involves labeling a few additional samples from the target domain to help transfer knowledge from the source domain to the target domain.

Existing active domain adaptation methods for semantic segmentation primarily rely on either modeling uncertainty~\cite{prabhu2021active, shin2021labor, xie2022towards} or data diversity~\cite{ning2021multi, xie2022towards, wu2022d} as metrics to select samples for annotation. They select regions with either higher uncertainty or larger differences from the source domain. However, these methods usually ignore the fact that selected samples can be highly correlated with their local context (e.g. local sample density and adjacent structures) in the feature space, which leads to inevitable redundancy within annotation budget.

\begin{figure}[t]
\centering
\includegraphics[width=0.6\linewidth]{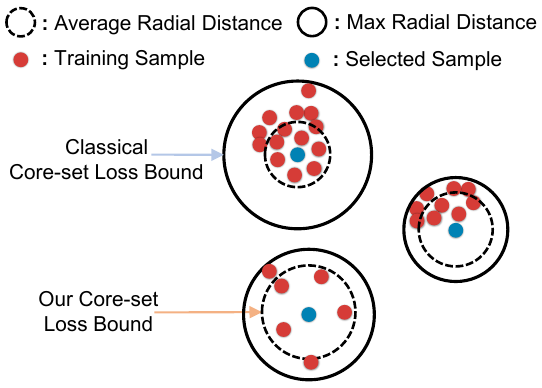}
\caption{Compared with the classical Core-set loss bound~\cite{sener2017active}, we find the `Average Radial Distance' in feature space is a tighter bound, which is highly correlated to local sample distribution.}
\label{fig:bound}
\end{figure}
One potential solution to the aforementioned problem is the Core-set approach, which uses a small set to approximate a large set~\cite{sener2017active, kim2022defense}. However, applying the Core-set approach to semantic segmentation with large amount of pixel-level candidates confronts two issues. \emph{Firstly}, the classical solution~\cite{wolf2011facility} to Core-set problems is essentially minimizing the radius required to cover extreme points in neighborhood (as shown in Fig.~\ref{fig:bound}). When the number of candidate points becomes larger, extreme points can not effectively reflect the local properties in feature space. \emph{Secondly}, in the classical greedy solution of Core-set, the discrepancy between data points is evaluated equally, regardless of the local context of each candidate. In contrast, the representative ability of samples may vary across different positions in the feature space, hence the discrepancy measurement should also depend on different data points.

To address these challenges, we first revisit the theoretical bound of classical Core-set loss~\cite{sener2017active} and derive a new upper bound from the perspective of expectation. This new bound indicates the Core-set performance is closely related to the conditional distribution of samples covered by its nearest selected data. More concretely, through intuitive analysis and empirical observation, we have discovered that the Core-set loss bound is scaled by the average distance from a sample to its closest selected points (as shown in Fig.~\ref{fig:bound}), which indicates that our selection strategy should allocate more label budget to samples with larger and more diverse coverage area. Consequently, we propose a Density-aware Core-set Selection method for domain adaptive semantic segmentation, which takes local sample distribution of candidates into account for domain adaptive semantic segmentation.

To estimate the average distance from a training sample to its nearest selected points, we draw inspiration from VAE~\cite{kingma2013auto} and the context model of learned image compression~\cite{minnen2018joint}, and design a fast local proxy estimator equipped with Dynamic Masked Convolution. This estimator estimates a statistic we call `coverage density' based on each candidate's neighbors, which is negatively related to the average distance. Additionally, we also develop a Density-aware Greedy algorithm that considers both data discrepancy and estimated coverage density when selecting samples, and minimizes the average radial distance within a defined coverage area of selected samples.

Overall, our contributions can be summarized as follows:
\begin{itemize}
    \item We derive a new upper bound for the Core-set approach based on an expectation perspective. Our analysis reveals that the average distances within a defined coverage area of selected points are crucial for the performance, which is highly correlated to the conditional probability density within the area.
    \item We propose a Density-aware Core-set Selection method to optimize the derived upper bound. We applies a proxy estimator equipped with Dynamic Mask Convolution for fast estimation of local density and average distance, the estimated density is then utilized in our proposed Density-aware Greedy algorithm for data sampling.
    \item We conduct experiments on two representative domain adaptation benchmarks, namely GTAV $\rightarrow$ Cityscapes and SYNTHIA $\rightarrow$ Cityscapes, achieving performance that surpasses current state-of-the-art methods.
\end{itemize}

\section{Related Works}
\subsection{Domain Adaptation}
Domain adaptation aims to transfer knowledge from a source domain with sufficient labels to a related target domain with little or no labeled data. Depending on the availability of labels in the target domain, domain adaptation can be classified into unsupervised domain adaptation (UDA) and weakly-supervised domain adaptation (WDA). Lately, UDA methods including HRDA~\cite{hoyer2022hrda} and MIC~\cite{hoyer2023mic} have achieved promising performance by designing efficient self-training strategies~\cite{wang2021uncertainty, zheng2021rectifying, jiang2022prototypical}.
As a complement to these works, this paper focuses on the WDA setting, which balances model performance and annotation cost through the use of weak labels such as image-level annotations~\cite{paul2020domain} or a limited number of pixel-wise labels~\cite{chen2021semi, guan2023iterative}.

\subsection{Active Domain Adaptation Segmentation}
Active learning is a powerful technique that improves model performance with a fixed labeling budget by selecting valuable samples for labeling in multiple rounds. 
Many active learning works focus on image classification~\cite{wang2016cost, xie2022learning, kirsch2019batchbald}. However, these methods are often not applicable to semantic segmentation with numerous candidate samples. For example, BADGE~\cite{ash2019deep} and BAIT~\cite{ash2021gone} incorporates last-layer gradients or Fisher matrices for active selection, but computing them for each pixel or image patch is computationally and storage-intensive.

Active domain adaptation methods have been developed for semantic segmentation to address the high cost of annotation and the performance of trained models. They typically employ unsupervised domain adaptation to initialize a model and subsequently choose samples in the target domain using indicators such as uncertainty~\cite{shin2021labor} or diversity~\cite{ning2021multi}. Some recent works have integrated uncertainty and diversity measures. RIPU~\cite{xie2022towards} uses the entropy of the model output to quantify uncertainty and measures diversity based on the number of classes predicted by the model within a fixed neighborhood. D2ADA \cite{wu2022d} utilizes the KL divergence of the feature distribution between the target and source domains to measure diversity. Nevertheless, these methods do not consider that selecting highly similar samples may result in a wasted labeling budget.

\subsection{Core-set Approach}
The Core-set selects a subset of data that approximates the entire dataset by choosing samples that cover the entire training set with the smallest possible radius, thereby enhancing the diversity of selected samples. Sener et al.~\cite{sener2017active} extended this approach to convolutional neural networks and developed a Robust k-Center algorithm to improve its optimality. Kim et al.~\cite{kim2022defense} first clustered samples in the training set according to the estimated nearest neighbor distance and then performed active selection in each cluster using the K-Center Greedy algorithm for image classification. However, this approach still relies on extreme data points in selection, ignoring the relationship between data discrepancy and local context.

In this paper, we derived a tighter upper bound for the Core-set loss and optimized it by assigning varying coverage radii to different samples in the selected set. To the best of our knowledge, this is the first time that the Core-set approach has been applied to semantic segmentation.

\section{Problem Definition with a Tighter Core-set Upper Bound}
First, we provide a brief overview of the optimization target of the Core-set approach and introduce some notation. The data space is denoted as $\mathcal X$ and the label space is denoted as $\mathcal Y = \{ 1, ..., C \}$, where $C$ represents the number of classes for semantic segmentation. The training set comprises independently and identically distributed (i.i.d.) samples from the space $\mathcal Z = \mathcal X \times \mathcal Y$. It is represented as $\{ \mathbf{x}_t, y_t \}_{t\in [n]}\sim p_{\mathcal Z}$, where $n$ is the size of the training set and $[n]$ is the set of subscripts $\{1,2,\cdots,n\}$. The Core-set approach aims to select a small subset of labeled samples $\mathbf{s}\subset [n]$ that minimizes the Core-set loss:
\begin{align}
    \min_{\mathbf{s}}\mathcal{L}(\mathbf{s}) = |\frac{1}{n}\sum_{t=1}^n & l(\mathbf{x}_t, y_t;A_{\mathbf{s}})-\frac{1}{|\mathbf{s}|}\sum_{k\in \mathbf{s}}l(\mathbf{x}_k, y_k; A_{\mathbf{s}})|, \notag \\
    & \textbf{s.t.} \quad |\mathbf{s}| = b,
\label{equ:core_set_target}
\end{align}
where $b$ is the labeling budget, $A_{\textbf{s}}$ represents learning algorithm which fits the class-wise distribution $\eta_c(\mathbf{x})=p(y=c|\mathbf{x})$ for each class given labeled subset $\mathbf{s}$, and $l(\cdot, \cdot, A_s)$ denotes a bounded non-negative loss function. Intuitively, Eq.~\ref{equ:core_set_target} aims at finding the subset $\mathbf{s}$ such that the performance of the model on $\mathbf{s}$ is close to that on all sampled data.

Previous research~\cite{sener2017active} demonstrated that, under appropriate assumptions, the objective of Eq.~\ref{equ:core_set_target} is bounded by the radius of balls determined by extreme points in data space.
\begin{theorem}\label{theorem:sener}
    Classical Core-set loss bound~\cite{sener2017active}: Given selected set s, if $l(\cdot, y, A_{\mathbf{s}})$ is $\lambda^l$-Lipschitz continuous and bounded by $L$, $\eta_c(\mathbf{x})$ is $\lambda^{\eta}$-Lipschitz continuous, and $l(\mathbf{x}_k, y_k, A_{\mathbf{s}})=0, \forall k\in \mathbf{s}$, then there exists a radius $\delta = \max_{\mathbf{x} \in \mathcal{X}}\min_{k \in \mathbf{s}}{|\mathbf{x} - \mathbf{x}_k|}$ such that, with probability no less than $1-\gamma$,
    \begin{align}
    \mathcal{L}(\mathbf{s}) \leq \delta (\lambda^l+\lambda^{\eta}LC) + \sqrt{\frac{L^2 log(1/ \gamma)}{2n}}.
    \label{equ:last_upper_bound}
    \end{align}
\end{theorem}

According to Eq.~\ref{equ:last_upper_bound}, minimizing the upper bound of the Core-set loss requires finding a subset $\mathbf{s}$ with the minimum required radius to cover all other samples, which in turn minimizes $\delta$. One approach is to use the k-Center Greedy algorithm~\cite{wolf2011facility} to obtain a sub-optimal solution. Alternatively, the radius $\delta$ can be reformulated as follows:
\begin{align}
\label{equ:coverage_area}
    & \delta = \max_k \max_{\mathbf{x} \in \mathbb{N}_{\mathbf{s}}(k)}|\mathbf{x} - \mathbf{x}_k|, \notag \\
    \mathbb{N}_{\mathbf{s}}(k)=\{\mathbf{x}| & \mathbf{x} \in \mathcal{X} \wedge \arg\min_{m \in \mathbf{s}}| \mathbf{x}-\mathbf{x}_m|=k\},
\end{align}
where we define $\mathbb{N}_{\mathbf{s}}(k)$ as the ``Coverage Area'' of points $\mathbf{x}_k, k \in \mathbf{s}$. As shown in Eq.~\ref{equ:coverage_area}, the upper bound defined by~\cite{sener2017active} only considers the furthest points (the inner max operation of Eq.~\ref{equ:coverage_area}) covered by each sample from $\mathbf{s}$, making it a relatively loose upper bound. In contrast, we claim that the Core-set loss can be bounded via a new formulation from the expectation view.

\begin{theorem}
With the same assumption as Theorem~\ref{theorem:sener} and definition of Coverage Area in Eq.~\ref{equ:coverage_area}, with probability at least $1-\gamma$, the Core-set loss is bounded by
\begin{align}
     \mathcal{L}(\mathbf{s}) \leq \max_{k\in \mathbf{s}} & \delta_{\mathbf{s}}(k) \cdot (\lambda^l+\lambda^{\eta}LC) + \sqrt{\frac{L^2 log(1/ \gamma)}{2n}}, \notag \\
     & \delta_{\mathbf{s}}(k) = \mathbb{E}_{\mathbf{x}\sim\mathbb{N}_{\mathbf{s}}(k)}[|\mathbf{x}-\mathbf{x}_k|].
\label{equ:tighter_upper_bound}
\end{align}
\label{theorem1}
\end{theorem}
Furthermore, it can be easily verified that the upper bound given by Theorem~\ref{theorem1} is a tighter approximation compared with the one in Theorem~\ref{theorem:sener}.
\begin{theorem}
With the same assumption as Theorem~\ref{theorem:sener} and definition of Coverage Area in Eq.~\ref{equ:coverage_area}, the upper-bound of Theorem~\ref{theorem1} is smaller than the classical Core-set bound~\cite{sener2017active}, i.e. $\max_{k\in\mathbf{s}}\delta_{\mathbf{s}}(k) \leq \delta$.
\label{theorem2}
\end{theorem}
Theorem~\ref{theorem1} shows that the bound is dependent on the maximum value of $\delta_{\mathbf{s}}(k)$, which represents the expected distance between $\mathbf{x}_k$ and other training samples lying in its coverage area $\mathbb{N}_{\mathbf{s}}(k)$. We define this factor $\delta_{\mathbf{s}}(k)$ as the "Average Radial Distance" of $\mathbf{x}_k, k \in \mathbf{s}$. Additionally, Theorem~\ref{theorem1} demonstrates that minimizing the maximum average radial distance of labeled set $\mathbf{s}$ is increasingly crucial for reducing the Core-set loss.

\begin{figure*}[t!]
\centering
\includegraphics[width=0.8\linewidth]{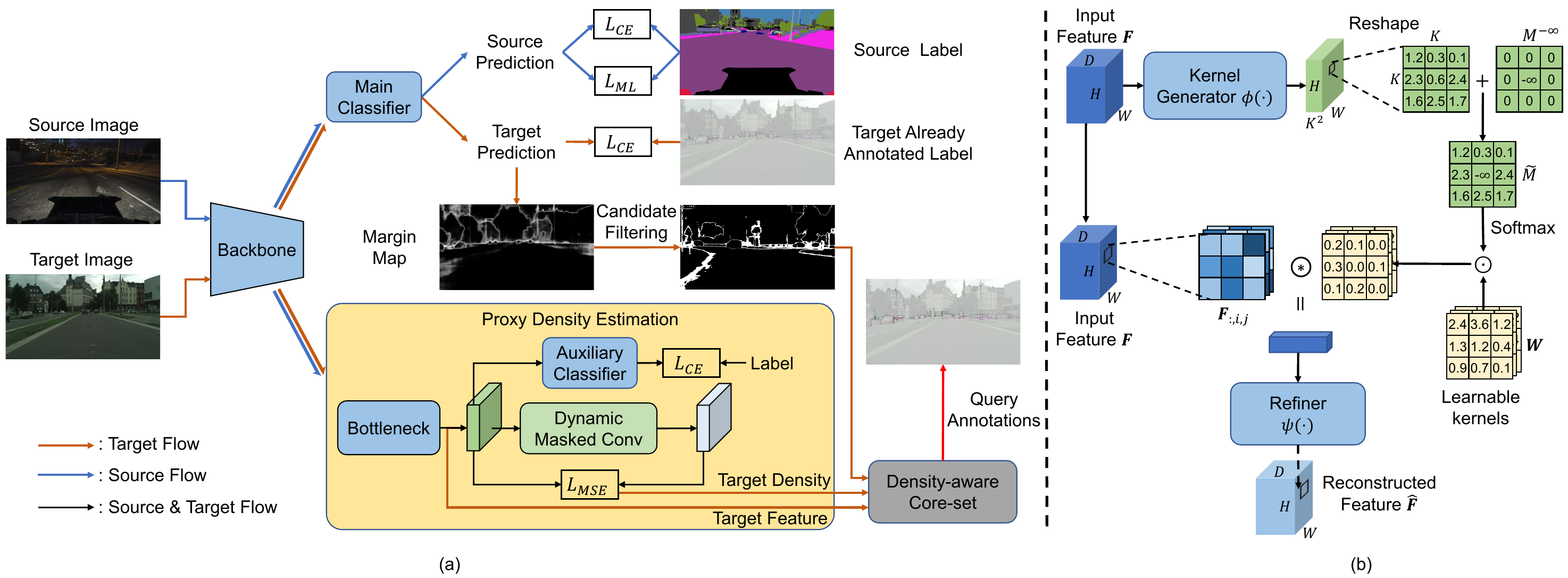}
\caption{(a) The overview of the proposed method. At each round of active selection, we first select $\alpha b (\alpha > 1)$ pixels closest to the classification boundary as candidate samples based source domain knowledge. We then introduce a density estimation branch to estimate the coverage densities of candidate samples. The features and densities of candidate samples are then fed into our proposed Density-aware Greedy algorithm to perform active selection. Finally, the network will be retrained using all available labels. (b) Structure of our proposed Dynamic Masked Convolution.}
\label{fig:framework}
\end{figure*}

\section{Density-aware Core-set with a Proxy Estimator}
\label{sec:method}
In this section we introduce how to minimize the upper-bound drawn from Theorem~\ref{theorem1} in the context of semantic segmentation. From the definition of average radial distance: 
\begin{align}
\label{equ:coverage_density}
    \delta_{\mathbf{s}}(k) = \int_{\mathbf x} |\mathbf{x}-\mathbf{x}_k| p (\mathbf x | \pi(\mathbf x)=k) d\mathbf x,
\end{align}
where $\pi(\mathbf{x}) = \arg\min_{k\in \mathbf{s}}|\mathbf{x} - \mathbf{x}_k|$.

We can observe that $\delta_{\mathbf{s}}(k)$ relies on the conditional probability distribution $p (\mathbf x | \pi(\mathbf x)=k)$ of samples within the coverage area $\mathbb{N}(\mathbf{x}_k)$ of the chosen sample $\mathbf{x}_k$. We term this distribution as 'coverage sample distribution' of $\mathbf{x}_k$. When samples are uniformly distributed in the feature space, the distributions of all samples are equal. As a consequence, the k-Center Greedy algorithm employed in \cite{sener2017active} effectively optimizes both the upper bounds in Eq. \ref{equ:last_upper_bound} and Eq. \ref{equ:tighter_upper_bound}. Nonetheless, samples in the feature space is seldom evenly distributed (One visual demonstration can be found in the supplementary). Consequently, a selected sample with more eccentric coverage sample distribution exhibit larger average radial distance, thereby it can exert greater influence on the bound of Eq.~\ref{equ:tighter_upper_bound}. Therefore, we drew inspiration to devise an optimization algorithm that takes the coverage sample distribution into consideration.

A naive idea is to modify the k-Center Greedy algorithm, i.e. at each step, instead of furthest point sampling, greedily select a sample that minimizes the maximum average radial distance of the newly selected set. However, such solution results in the time complexity of $\mathcal{O}(n^2b)$, which is almost unacceptable in the large candidate number scenario of semantic segmentation. Therefore, we propose a fast local proxy estimator to estimate a `coverage density' that reflects the average radial distance for each pixel. Subsequently, this coverage density is utilized in the Density-aware Greedy algorithm (shown in Fig. \ref{fig:framework}).

\subsection{Baseline Training and Candidate Selection}
In our framework, the images from the source and target domains are separately sent to the Backbone to extract features. The prediction $P \in R^{C\times H_I \times W_I}$ is obtained through a main classifier, where $H_I\times W_I$ represents the resolution of original image. All labeled data is used to train the backbone and main classifier via cross-entropy. 

\textbf{Source Aided Candidate Filtering.} In active semantic segmentation, data is selected at the pixel-level, resulting in extremely large search space for Core-set selection. Therefore, we preselect informative candidate points from the target data. In order to fully utilize the source domain knowledge to filter out informative pixels in the target domain that are distinct from the source domain, we applied the categorical-wise margin loss proposed in \cite{xie2022learning} to the source data, as shown in Eq.~\ref{equ:margin_loss}.
\begin{equation}
    \mathcal{L}_{ML} = \sum_i \sum_j \sum_{c\neq y} [m-P_{y, i, j}+P_{c, i, j}]_+
    \label{equ:margin_loss}
\end{equation}
where $m$ is set to 1 to control the margin width, $y$ corresponds to the channel index of the ground-truth, and $[x]_+$ denotes $max(0, x)$. Subsequently, we can select informative target samples based on the source knowledge by utilizing $I_{i,j}=1-P_{1*, i, j}+P_{2*, i, j}$, where $P_{1*, i, j}$ and $P_{2*, i, j}$ are respectively the maximum and second maximum values within $P_{:,i,j}$.
For annotation budget $b$, we select the top $\alpha b$ ($\alpha>1$) pixels with the highest $I_{i,j}$ as candidate samples to reduce complexity. These selected candidates are then resampled through the Density-aware Greedy algorithm based on their estimated density.

\subsection{Proxy Density Estimator with Dynamic Masked Convolution}
Due to the large size and high dimensionality of semantic segmentation candidates, classic distribution estimation methods such as GMM~\cite{reynolds2009gaussian} cannot estimate the coverage sample distribution accurately in an computationally efficient way. Therefore, we propose a proxy estimator for fast estimation. We employ the concept of the Monte Carlo method to estimate a statistic reflecting the local sample distribution of each pixel, termed the 'coverage density'. Given the continuity of images, spatially adjacent pixels tend to be also adjacent in the feature space~\cite{qian2022entroformer}. Consequently, the features of neighboring pixels can be considered as samples drawn from the central pixel's coverage sample distribution $p (\mathbf x | \pi(\mathbf x)=k)$. The process of calculating distances between the central pixel and its neighboring pixels and averaging them is equivalent to applying a Monte Carlo method to approximate Eq.~\ref{equ:coverage_density}. However, such naive estimation can results bias since the local spatial window cannot contain enough samples. To improve this estimation under limited neighboring pixels, we introduce the Dynamic Masked Convolution module (DMC), which aggregates neighboring pixel features to reconstruct the central pixel's feature. Since maximizing the likelihood of a Gaussian distribution is equivalent to minimizing the MSE~\cite{kingma2013auto}, our DMC aligns with the use of masked convolution in learned image compression~\cite{minnen2018joint} to estimate the pixel-wise local conditional distribution.

\begin{figure*}[t]
\begin{center}
\includegraphics[width=0.8\linewidth]{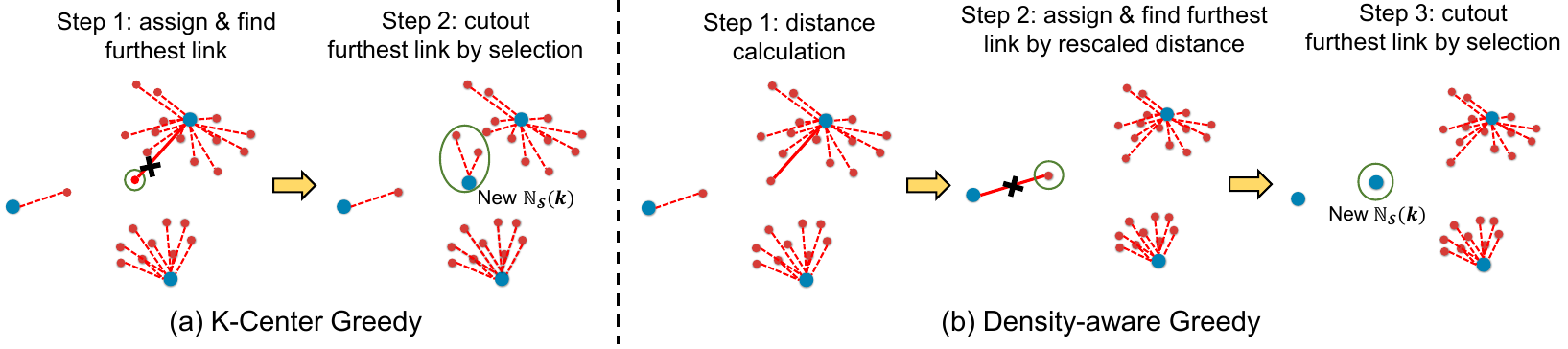}
\end{center}
\caption{Comparison between k-Center Greedy and Density-aware Greedy. The red points represent candidates to be selected, and the blue points denote selected samples in $\mathbf{s}$. (a) The two-step ``Find \& Cut'' view of k-Center Greedy selection, which cuts the link of furthest distance. (b) Our Density-aware Greedy algorithm rescales the distances so that selected samples with higher density are closer to other candidates and low-density ones are pushed away from linked candidates.}
\label{fig:scheme_comparison}
\end{figure*}

\begin{figure}[t]
    \centering
    \includegraphics[width=0.55\linewidth]{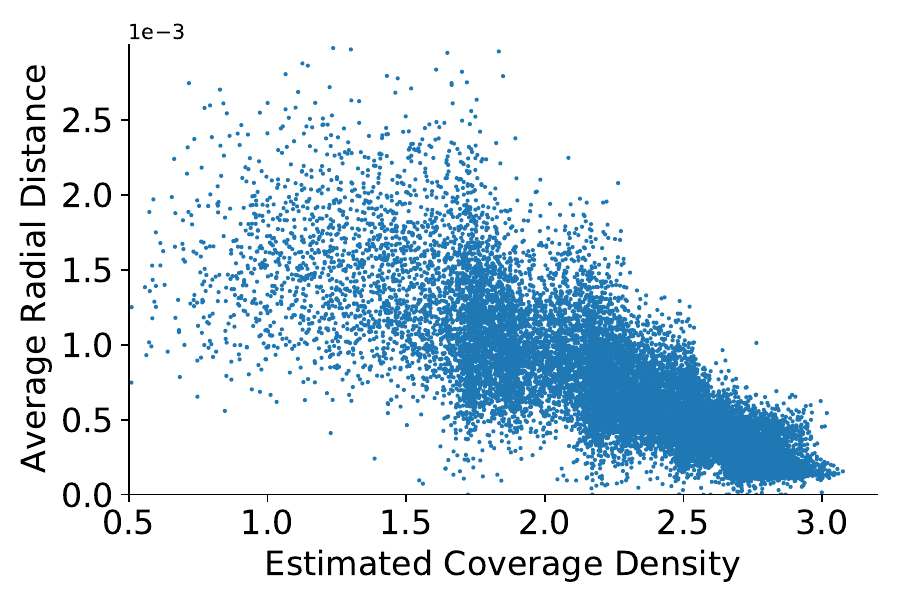}
    \caption{Experimental observation of the relation between estimated density and average radial distance. The distribution of average radial distance tends to approach zero w.r.t increasing coverage density.}
    \label{fig:exp_observe}
\end{figure}
To enable faster estimation, we first introduce a convolution layer $g(\cdot)$ to convert original feature from backbone to a low-dimensional representation $\mathbf{F}\in R^{D\times H\times W}$ with channel $D$ and spatial size $H\times W$. Next we stack several convolutions $h(\cdot)$ and a softmax operation $\sigma(\cdot)$ as dynamic mask generator $\phi(\cdot)$ to obtain the spatial modulation map $\mathcal{M}=\phi(\mathbf{F})=\sigma\left(h(\mathbf{F})+\mathcal{M}^{-\infty}\right) \in R^{K^2\times H\times W}$, where $K$ is the kernel size for dynamic convolution. Before the softmax $\sigma$, we add a mask tensor $\mathcal{M}^{-\infty} \in \{0, -\infty\}^{K^2\times H\times W}$, where the $\lceil \frac{K^2}{2}\rceil$-th channel is set $-\infty$ to mask out the weight of center coordinates in a $K\times K$ kernel. We then reshape $\mathcal{M}$ as $\tilde{\mathcal{M}}\in R^{K\times K\times H \times W}$, and the reconstructed representation $\tilde{\mathbf{F}}$ is expressed as:
{\scriptsize
\begin{align}
     \tilde{\mathbf{F}}_{o,i,j}= \sum_{d=1}^D \sum_{(u, v)\in \Delta K} \tilde{\mathcal{M}}_{u+\lfloor \frac{K}{2} \rfloor, v + \lfloor \frac{K}{2} \rfloor, i, j} \cdot \notag \\ 
     \mathbf{F}_{d,i+u,j+v} \cdot \mathbf{W}_{o, d, u +\lfloor \frac{K}{2} \rfloor, v + \lfloor \frac{K}{2} \rfloor}
\end{align}
}
where $\Delta K$ represents $[-\lfloor \frac{K}{2}\rfloor, \lfloor \frac{K}{2}\rfloor]\times [-\lfloor \frac{K}{2}\rfloor, \lfloor \frac{K}{2}\rfloor]$, and $\mathbf{W} \in R^{D\times D\times K\times K}$ is the learnable parameters of dynamic convolution. Further, additional convolution layers $\psi(\cdot)$ are used to refine the reconstructed results $\hat{\mathbf{F}}=\psi(\tilde{\mathbf{F}})$, where the convolution kernel in $\psi(\cdot)$ is $1\times 1$ to avoid spatial information leakage.
Finally, we exploit the reconstruction error to approximate the coverage density at location $(i,j)$ as $ \mathbf{D}_{i,j} = \beta\exp\left(-{||\hat{\mathbf{F}}_{:,i,j} - \mathbf{F}_{:,i,j}||_2^2/\tau}\right)$,
where both $\beta,\tau$ are hyperparameters, and we regard the inner reconstruction error $||\hat{\mathbf{F}}_{:,i,j} - \mathbf{F}_{:,i,j}||_2^2$ as estimated average radial distance. We train the estimator to extimate the coverage density for both labeled and unlabeled data:
\begin{equation}\label{equ:max_density}
    \max_{g, \phi, \psi}\log{\prod_{(i,j)}{\mathbf{D}_{i,j}}}
\end{equation}
Meanwhile, to prevent the optimization in Eq.~\ref{equ:max_density} from having a trivial solution, we append another auxiliary classifier on $\mathbf{F}$ and supervise it with cross-entropy loss using labeled data (as shown in Fig.~\ref{fig:framework}). Before feeding each candidate pixel into the following Density-aware Greedy algorithm, we extract its feature vector $\mathbf{f}$ and density $\mathbf{d}$ from $\mathbf{F}$ and $\mathbf{D}$ at corresponding spatial location respectively.

From Eq.~\ref{equ:max_density}, we can observe that the estimated coverage density is negative correlated with the average radial distance. To demonstrate this, in Fig.~\ref{fig:exp_observe}, we analyze the correlation between the estimated density and the average radial distance given the selected set $\mathbf{s}$ from the naive Core-set~\cite{sener2017active}. The average radial distance of labeled samples with low density is more likely to be larger, while as the density increases, the distribution of the average radial distance tends to approach zero.

\begin{table*}[t]
    \centering
    \caption{Comparison with various domain adaptation methods on GTAV $\rightarrow$ Cityscapes.}
    \label{tab:gtav2city}
    \begin{adjustbox}{width=\linewidth}
    \begin{tabular}{l|ccccccccccccccccccc|c}
        \toprule
        \makecell*[c]{Method} & \makecell*[c]{Road} & \makecell*[c]{SW} & \makecell*[c]{Build} & \makecell*[c]{Wall} & \makecell*[c]{Fence} & \makecell*[c]{Pole} & \makecell*[c]{Light} & \makecell*[c]{Sign} & \makecell*[c]{Veg.} & \makecell*[c]{Terr.} & \makecell*[c]{Sky} & \makecell*[c]{Pers.} & \makecell*[c]{Rider} & \makecell*[c]{Car} & \makecell*[c]{Truck} & \makecell*[c]{Bus} & \makecell*[c]{Train} & \makecell*[c]{MB} & \makecell*[c]{Bike} & \makecell*[c]{mIOU} \\
        \cmidrule(lr){1-1} \cmidrule(lr){2-20} \cmidrule(lr){21-21}
        Source Only & 75.8 & 16.8 & 77.2 & 12.5 & 21.0 & 25.5 & 30.1 & 20.1 & 81.3 & 24.6 & 70.3 & 53.8 & 26.4 & 49.9 & 17.2 & 25.9 & 6.5 & 25.3 & 36.0 & 36.6\\
        AdaptSeg \cite{tsai2018learning} & 86.5 & 36.0 & 79.9 & 23.4 & 23.3 & 35.2 & 14.8 & 14.8 & 83.4 & 33.3 & 75.6 & 58.5 & 27.6 & 73.7 & 32.5 & 35.4 & 3.9 & 30.1 & 28.1 & 42.4\\
        CBST \cite{zou2018unsupervised} & 91.8 & 53.5 & 80.5 & 32.7 & 21.0 & 34.0 & 28.9 & 20.4 & 83.9 & 34.2 & 80.9 & 53.1 & 24.0 & 82.7 & 30.3 & 35.9 & 16.0 & 25.9 & 42.8 & 45.9\\
        CRST \cite{zou2019confidence} & 91.0 & 55.4 & 80.0 & 33.7 & 21.4 & 37.3 & 32.9 & 24.5 & 85.0 & 34.1 & 80.8 & 57.7 & 24.6 & 84.1 & 27.8 & 30.1 & 26.9 & 26.0 & 42.3 & 47.1\\
        FDA \cite{yang2020fda} & 92.5 & 53.3 & 82.4 & 26.5 & 27.6 & 36.4 & 40.6 & 38.9 & 82.3 & 39.8 & 78.0 & 62.6 & 34.4 & 84.9 & 34.1 & 53.1 & 16.9 & 27.7 & 46.4 & 50.5\\
        TPLD \cite{shin2020two} & 94.2 & 60.5 & 82.8 & 36.6 & 16.6 & 39.3 & 29.0 & 25.5 & 85.6 & 44.9 & 84.4 & 60.6 & 27.4 & 84.1 & 37.0 & 47.0 & 31.2 & 36.1 & 50.3 & 51.2\\
        IAST \cite{mei2020instance} & 93.8 & 57.8 & 85.1 & 39.5 & 26.7 & 26.2 & 43.1 & 34.7 & 84.9 & 32.9 & 88.0 & 62.6 & 29.0 & 87.3 & 39.2 & 49.6 & 23.2 & 34.7 & 39.6 & 51.5\\
        DPL-Dual \cite{cheng2021dual} & 92.8 & 54.4 & 86.2 & 41.6 & 32.7 & 36.4 & 49.0 & 34.0 & 85.8 & 41.3 & 86.0 & 63.2 & 34.2 & 87.2 & 39.3 & 44.5 & 18.7 & 42.6 & 43.1 & 53.3\\
        BAPA-Net \cite{liu2021bapa} & 94.4 & 61.0 & 88.0 & 26.8 & 39.9 & 38.3 & 46.1 & 55.3 & 87.8 & 46.1 & 89.4 & 68.8 & 40.0 & 90.2 & 60.4 & 59.0 & 0.0 & 45.1 & 54.2 & 57.4\\
        ProDA \cite{zhang2021prototypical} & 87.8 & 56.0 & 79.7 & 46.3 & 44.8 & 45.6 & 53.5 & 53.5 & 88.6 & 45.2 & 82.1 & 70.7 & 39.2 & 88.8 & 45.5 & 59.4 & 1.0 & 48.9 & 56.4 & 57.5\\
        \cmidrule(lr){1-21}
        WeakDA (point) \cite{paul2020domain} & 94.0 & 62.7 & 86.3 & 36.5 & 32.8 & 38.4 & 44.9 & 51.0 & 86.1 & 43.4 & 87.7 & 66.4 & 36.5 & 87.9 & 44.1 & 58.8 & 23.2 & 35.6 & 55.9 & 56.4\\
        \cmidrule(lr){1-21}
        LabOR (V2, 40 pixels) \cite{shin2021labor} & 96.1 & 71.8 & \textbf{88.8} & 47.0 & 46.5 & \textbf{42.2} & \textbf{53.1} & 60.6 & \textbf{89.4} & 55.1 & \textbf{91.4} & 70.8 & 44.7 & 90.6 & 56.7 & 47.9 & 39.1 & 47.3 & 62.7 & 63.5\\
        RIPU (V2, 40 pixels) \cite{xie2022towards} & 95.5 & 69.2 & 88.2 & 48.0 & \textbf{46.5} & 36.9 & 45.2 & 55.7 & 88.5 & \textbf{55.3} & 90.2 & 69.2 & 46.1 & 91.2 & 70.7 & \textbf{73.0} & \textbf{58.2} & 50.1 & 65.9 & 65.5\\
        Ours (V2, 40 pixels) & \textbf{96.4} & \textbf{73.4} & 88.7 & \textbf{49.9} & 44.1 & 40.7 & 49.6 & \textbf{60.7} & 89.3 & 54.9 & 91.2 & \textbf{71.0} & \textbf{48.7} & \textbf{91.6} & \textbf{71.9} & 71.8 & 53.9 & \textbf{55.3} & \textbf{68.3} & \textbf{66.9}\\
        \cmidrule(lr){1-21}
        LabOR (V2, 2.2\%) \cite{shin2021labor} & 96.6 & 77.0 & 89.6 & 47.8 & 50.7 & \textbf{48.0} & 56.6 & 63.5 & 89.5 & 57.8 & 91.6 & 72.0 & 47.3 & 91.7 & 62.1 & 61.9 & 48.9 & 47.9 & 65.3 & 66.6\\
        RIPU (V2, 2.2\%) \cite{xie2022towards} & 96.5 & 74.1 & 89.7 & 53.1 & 51.0 & 43.8 & 53.4 & 62.2 & 90.0 & 57.6 & 92.6 & 73.0 & 53.0 & 92.8 & 73.8 & 78.5 & 62.0 & 55.6 & 70.0 & 69.6\\
        D2ADA (V2, 5\%) \cite{wu2022d} & 96.3 & 73.6 & 89.3 & 50.0 & \textbf{52.3} & 48.0 & \textbf{56.9} & 64.7 & 89.3 & 53.9 & 92.3 & 73.9 & 52.9 & 91.8 & 69.7 & \textbf{78.9} & 62.7 & \textbf{57.7} & \textbf{71.1} & 69.7\\
        \textbf{Ours (V2, 2.2\%)} & \textbf{97.4} & \textbf{79.4} & \textbf{90.4} & \textbf{56.9} & 51.5 & 45.8 & 56.6 & \textbf{67.2} & \textbf{90.3} & \textbf{58.5} & \textbf{92.9} & \textbf{74.2} & \textbf{55.0} & \textbf{92.8} & \textbf{75.8} & 75.0 & \textbf{65.3} & 54.5 & 70.4 & \textbf{71.1}\\
        Fully Supervised (V2) & 97.4 & 79.5 & 90.3 & 51.1 & 52.4 & 49.0 & 57.5 & 68.0 & 90.5 & 58.1 & 93.1 & 75.1 & 53.9 & 92.7 & 72.0 & 80.2 & 65.0 & 58.1 & 71.1 & 71.3\\
        \cmidrule(lr){1-21}
        MADA (V3+, 5\%) \cite{ning2021multi} & 95.1 & 69.8 & 88.5 & 43.3 & 48.7 & 45.7 & 53.3 & 59.2 & 89.1 & 46.7 & 91.5 & 73.9 & 50.1 & 91.2 & 60.6 & 56.9 & 48.4 & 51.6 & 68.7 & 64.9\\
        RIPU (V3+, 5\%) \cite{xie2022towards} & 97.0 & 77.3 & 90.4 & \textbf{54.6} & 53.2 & 47.7 & 55.9 & 64.1 & 90.2 & 59.2 & 93.2 & 75.0 & 54.8 & 92.7 & \textbf{73.0} & \textbf{79.7} & 68.9 & 55.5 & 70.3 & 71.2\\
        D2ADA (V3+, 5\%) \cite{wu2022d} & 97.0 & 77.8 & 90.0 & 46.0 & \textbf{55.0} & 52.7 & \textbf{58.7} & 65.8 & 90.4 & 58.9 & 92.1 & 75.7 & 54.4 & 92.3 & 69.0 & 78.0 & 68.5 & \textbf{59.1} & \textbf{72.3} & 71.3\\
        \textbf{Ours (V3+, 5\%)} & \textbf{97.6} & \textbf{81.1} & \textbf{90.9} & 52.7 & 52.8 & \textbf{53.9} & 58.5 & \textbf{69.0} & \textbf{91.1} & \textbf{62.5} & \textbf{93.4} & \textbf{75.9} & \textbf{54.8} & \textbf{92.9} & 72.5 & 76.5 & \textbf{71.3} & 54.2 & 71.2 & \textbf{72.2}\\
        Fully Supervised (V3+) & 97.6 & 81.3 & 91.1 & 49.8 & 57.6 & 53.8 & 59.6 & 69.1 & 91.2 & 60.5 & 94.4 & 76.7 & 55.6 & 93.3 & 75.8 & 79.9 & 72.9 & 57.7 & 72.2 & 73.2\\
        \bottomrule
        \end{tabular}
\end{adjustbox}
\end{table*}
\begin{table*}[t]
    \centering
    \caption{Comparison with various domain adaptation methods on SYNTHIA $\rightarrow$ Cityscapes.}
    \label{tab:synthia2city}
    \begin{adjustbox}{width=\linewidth}
    \begin{tabular}{l|cccccccccccccccc|cc}
    \toprule
         \makecell*[c]{Method} & \makecell*[c]{Road} & \makecell*[c]{SW} & \makecell*[c]{Build} & \makecell*[c]{Wall*} & \makecell*[c]{Fence*} & \makecell*[c]{Pole*} & \makecell*[c]{Light} & \makecell*[c]{Sign} & \makecell*[c]{Veg.} & \makecell*[c]{Sky} & \makecell*[c]{Pers.} & \makecell*[c]{Rider} & \makecell*[c]{Car} & \makecell*[c]{Bus} & \makecell*[c]{MB} & \makecell*[c]{Bike} & \makecell*[c]{mIOU} & \makecell*[c]{mIOU*}\\
         \cmidrule(lr){1-1} \cmidrule(lr){2-3} \cmidrule(lr){4-17}\cmidrule(lr){18-19}
        Source Only & 64.3 & 21.3 & 73.1 & 2.4 & 1.1 & 31.4 & 7.0 & 27.7 & 63.1 & 67.6 & 42.2 & 19.9 & 73.1 & 15.3 & 10.5 & 38.9 & 34.9 & 40.3\\
        AdaptSeg \cite{tsai2018learning} & 79.2 & 37.2 & 78.8 & - & - & - & 9.9 & 10.5 & 78.2 & 80.5 & 53.5 & 19.6 & 67.0 & 29.5 & 21.6 & 31.3 & - & 45.6 \\
        CBST \cite{zou2018unsupervised} & 68.0 & 29.9 & 76.3 & 10.8 & 1.4 & 33.9 & 22.8 & 29.5 & 77.6 & 78.3 & 60.6 & 28.3 & 81.6 & 23.5 & 18.8 & 39.8 & 42.6 & 48.9 \\
        CRST \cite{zou2019confidence} & 67.7 & 32.2 & 73.9 & 10.7 & 1.6 & 37.4 & 22.2 & 31.2 & 80.8 & 80.5 & 60.8 & 29.1 & 82.8 & 25.0 & 19.4 & 45.3 & 43.8 & 50.1 \\
        FDA \cite{yang2020fda} & 79.3 & 35.0 & 73.2 & - & - & - & 19.9 & 24.0 & 61.7 & 82.6 & 61.4 & 31.1 & 83.9 & 40.8 & 38.4 & 51.1 & - & 52.5 \\
        TPLD \cite{shin2020two} & 80.9 & 44.3 & 82.2 & 19.9 & 0.3 & 40.6 & 20.5 & 30.1 & 77.2 & 80.9 & 60.6 & 25.5 & 84.8 & 41.4 & 24.7 & 43.7 & 47.3 & 53.5 \\
        IAST \cite{mei2020instance} & 81.9 & 41.5 & 83.3 & 17.7 & 4.6 & 32.3 & 30.9 & 28.8 & 83.4 & 85.0 & 65.5 & 30.8 & 86.5 & 38.2 & 33.1 & 52.7 & 49.8 & 57.0 \\
        DPL-Dual \cite{cheng2021dual} & 87.5 & 45.7 & 82.8 & 13.3 & 0.6 & 33.2 & 22.0 & 20.1 & 83.1 & 86.0 & 56.6 & 21.9 & 83.1 & 40.3 & 29.8 & 45.7 & 47.0 & 54.2 \\
        BAPA-Net \cite{liu2021bapa} & 91.7 & 53.8 & 83.9 & 22.4 & 0.8 & 34.9 & 30.5 & 42.8 & 86.6 & 88.2 & 66.0 & 34.1 & 86.6 & 51.3 & 29.4 & 50.5 & 53.3 & 61.2 \\
        ProDA \cite{zhang2021prototypical} & 87.8 & 45.7 & 84.6 & 37.1 & 0.6 & 44.0 & 54.6 & 37.0 & 88.1 & 84.4 & 74.2 & 24.3 & 88.2 & 51.1 & 40.5 & 45.6 & 55.5 & 62.0\\
        \cmidrule(lr){1-19}
        WeakDA (point) \cite{paul2020domain} & 94.9 & 63.2 & 85.0 & 27.3 & 24.2 & 34.9 & 37.3 & 50.8 & 84.4 & 88.2 & 60.6 & 36.3 & 86.4 & 43.2 & 36.5 & 61.3 & 57.2 & 63.7 \\
        \cmidrule(lr){1-19}
        RIPU (V2, 40 pixels) \cite{xie2022towards} & 95.8 & 71.9 & 87.8 & 39.9 & 41.5 & 38.3 & 47.1 & 54.2 & 89.2 & 90.8 & 69.9 & \textbf{48.5} & 91.4 & 71.5 & \textbf{52.2} & 67.2 & 66.1 & 72.1 \\
        Ours (V2, 40 pixels) & \textbf{96.5} & \textbf{74.8} & \textbf{89.0} & \textbf{45.2} & \textbf{44.0} & \textbf{41.9} & \textbf{50.6} & \textbf{60.9} & \textbf{89.9} & \textbf{91.8} & \textbf{71.7} & 46.5 & \textbf{92.1} & \textbf{76.2} & 47.1 & \textbf{68.0} & \textbf{67.9} & \textbf{73.5}\\
        \cmidrule(lr){1-19}
        RIPU (V2, 2.2\%) \cite{xie2022towards} & 96.8 & 76.6 & 89.6 & 45.0 & 47.7 & 45.0 & 53.0 & 62.5 & 90.6 & 92.7 & 73.0 & 52.9 & 93.1 & 80.5 & 52.4 & 70.1 & 70.1 & 75.7 \\
        D2ADA (V2, 5\%) \cite{wu2022d} & 96.4 & 76.3 & 89.1 & 42.5 & 47.7 & \textbf{48.0} & 55.6 & 66.5 & 89.5 & 91.7 & \textbf{75.1} & \textbf{55.2} & 91.4 & 77.0 & \textbf{58.0} & \textbf{71.8} & 70.6 & 76.3\\
        \textbf{Ours (V2, 2.2\%)} & \textbf{97.4} & \textbf{80.9} & \textbf{90.4} & \textbf{48.4} & \textbf{51.5} & 47.7 & \textbf{56.5} & \textbf{68.0} & \textbf{91.2} & \textbf{93.0} & 74.8 & 52.2 & \textbf{93.4} & \textbf{83.5} & 54.6 & 70.7 & \textbf{72.1} & \textbf{77.3} \\
        Fully Supervised (V2) & 97.4 & 79.5 & 90.3 & 51.1 & 52.4 & 49.0 & 57.5 & 68.0 & 90.5 & 93.1 & 75.1 & 53.9 & 92.7 & 80.2 & 58.1 & 71.1 & 72.5 & 77.5 \\
        \cmidrule(lr){1-19}
        MADA (V3+, 5\%) \cite{ning2021multi} & 96.5 & 74.6 & 88.8 & 45.9 & 43.8 & 46.7 & 52.4 & 60.5 & 89.7 & 92.2 & 74.1 & 51.2 & 90.9 & 60.3 & 52.4 & 69.4 & 68.1 & 73.3 \\
        RIPU (V3+, 5\%) \cite{xie2022towards} & 97.0 & 78.9 & 89.9 & 47.2 & 50.7 & 48.5 & 55.2 & 63.9 & 91.1 & 93.0 & 74.4 & 54.1 & 92.9 & 79.9 & 55.3 & 71.0 & 71.4 & 76.7 \\
        D2ADA (V3+, 5\%) \cite{wu2022d} & 96.7 & 76.8 & 90.3 & \textbf{48.7} & 51.1 & 54.2 & 58.3 & 68.0 & 90.4 & 93.4 & \textbf{77.4} & \textbf{56.4} & 92.5 & 77.5 & \textbf{58.9} & \textbf{73.3} & 72.7 & 77.7\\
        \textbf{Ours (V3+, 5\%)} & \textbf{97.6} & \textbf{82.3} & \textbf{91.1} & 44.7 & \textbf{51.9} & \textbf{55.1} & \textbf{59.1} & \textbf{70.0} & \textbf{91.9} & \textbf{93.8} & 77.3 & 54.4 & \textbf{93.9} & \textbf{80.3} & 56.4 & 71.9 & \textbf{73.2} & \textbf{78.5} \\
        Fully Supervised (V3+) & 97.6 & 81.3 & 91.1 & 49.8 & 57.6 & 53.8 & 59.6 & 69.1 & 91.2 & 94.4 & 76.7 & 55.6 & 93.3 & 79.9 & 57.7 & 72.2 & 73.8 & 78.4 \\
        \bottomrule
        \end{tabular}
\end{adjustbox}
\end{table*}
\subsection{Density-aware Greedy Algorithm}
\label{sec:ada_radius_core_set}
\begin{algorithm}[t]
\caption{Density-aware Greedy}
\label{algo:ar_core_set}
\textbf{Input:} candidates $\mathbf{x}_t$, feature $\mathbf{f}_t$ and densities $\mathbf{d}_t$, $t\in [n]$, existing labeled set $\mathbf{s}^0$ and budget $b$

$\mathbf{s}=\mathbf{s}^0$ \\ $r_t = \min_{k\in \mathbf{s}} ||\mathbf{f}_t - \mathbf{f}_k||^2_2/\mathbf{d}_k \quad \forall t \in [n]$ \\
\textbf{repeat}

    \quad $u = \arg\max_{t\in [n] \backslash \mathbf{s}} r_t$

    \quad $\mathbf{s}=\mathbf{s}\cup \{ u \}$

    \quad $r_t = \min(r_t, ||\mathbf{f}_t - \mathbf{f}_u||^2_2/\mathbf{d}_u) \quad \forall t\in [n]\backslash\mathbf{s}$

\textbf{until} $|\mathbf{s}|=b + |\mathbf{s}^0|$

\textbf{return} $\mathbf{s}$
\end{algorithm}

With the estimated density, we propose a density-aware modification of k-Center Greedy algorithm~\cite{wolf2011facility} to minimize the Core-set upper bound. To make analogy, we first breakdown the k-Center algorithm into a two-step ``Find \& Cut'' manner as Fig.~\ref{fig:scheme_comparison}: \textbf{STEP1:} link $\mathbf{x}_t$ to its nearest point in $\mathbf{s}$ thus to obtain estimated $\mathbb{N}_{\mathbf{s}}(k)$ for $\mathbf{x}_k, \forall k \in \mathbf{s}$. \textbf{STEP2}: To shrink coverage of $\mathbb{N}_{\mathbf{s}}(k)$, find and cut the link with longest distance by appending corresponding candidate into $\mathbf{s}$. In contrast, our goal is to downgrade $\delta_{\mathbf{s}}(k)$ instead of shrinking $\mathbb{N}_{\mathbf{s}}(k)$, therefore we insert a step between \textbf{STEP1} and \textbf{STEP2} to \textbf{rescale the distance of links in each $\mathbb{N}_{\mathbf{s}}(k)$ by density $\mathbf{d}_k$ of corresponding labeled data}. When the coverage density estimated by DMC exhibits a strong correlation with the average radial distance, this inserted step transforms the cut link from furthest link in the feature space into the link associated with the maximum $\delta_{\mathbf{s}}$. As a result, selected samples with larger $\delta_{\mathbf{s}}$ are assigned smaller coverage areas and the maximum average radial distance is optimized. In practice, to ensure a robust correlation between the estimated coverage density and the average radial distance, we tune $\beta$ and $\tau$ over the training dataset until they reduce the bound in Eq.~\ref{equ:tighter_upper_bound}. Following~\cite{xie2022towards, wu2022d}, we perform multiple rounds of active selection. Algorithm~\ref{algo:ar_core_set} depicts one round of our proposed method.

\section{Experiments}
\subsection{Experimental Setup}
\textbf{Datasets.}
We evaluate our approach using two popular domain adaptive semantic segmentation benchmarks: GTAV $\rightarrow$ Cityscapes and Synthia $\rightarrow$ Cityscapes. 
GTAV and SYNTHIA are both synthetic datasets. GTAV shares 19 semantic categories with Cityscapes while SYNTHIA shares 16 semantic categories with Cityscapes.

\textbf{Implementation Details.}
To fairly compare with other methods, 
our training settings are aligned with ~\cite{xie2022towards}. 
In active selection, $\beta$ is set to $e^{2.4}$ and $\tau$ is set to $0.25$, normalizing the reconstruction error to 0-1 before calculating the density. The candidate features $\mathbf{f}$ are also normalized before being fed into Density-aware Greedy. We set $\alpha$ in the candidate filtering to 20 for label budgets lower than 2.2\%, and 10 for 5\% label budget.


\textbf{Active Learning Protocal.}
Similar to ~\cite{xie2022towards}, we performed 5 rounds of active selection, at 10000, 12000, 14000, 16000 and 18000 iterations.

\subsection{Comparison with State-of-the-Art Methods}
We compare our method with various domain adaptation methods, as shown in Table~\ref{tab:gtav2city} and ~\ref{tab:synthia2city}. Among them, ~\cite{tsai2018learning, zou2018unsupervised, zou2019confidence, yang2020fda, shin2020two, mei2020instance, cheng2021dual, liu2021bapa, zhang2021prototypical} are UDA methods, ~\cite{paul2020domain} is a WDA method, while ~\cite{shin2021labor, ning2021multi, xie2022towards, wu2022d} are ADA methods.

It can be observed that: (1) Compared with UDA and WDA methods, our method can achieve more than 10 mIOU improvement, which proves the effectiveness of active learning for domain adaptive semantic segmentation; (2) Compared with other active domain adaptation methods, our proposed method achieves significant improvements. Specifically, In comparison to RIPU, our method showcases a remarkable improvement of over 1.4 mIOU at labeling budgets of 40 pixels and 2.2\%. In comparison to D2ADA with Deeplab V2 and 5\% labeling budget, our approach achieves a boost of over 1.4 mIOU with a mere 2.2\% labeling budget. As the segmentation head improves and the labeling budget increases, the performance improvement brought by active selection strategy somewhat saturates. When we employ DeeplabV3+ at a labeling ratio of 5\%, our method outperforms RIPU by 1.0 mIOU and 1.8 mIOU respectively. In comparison to D2ADA, our method yields a performance increase of 0.9 mIOU and 0.5 mIOU respectively. (3) Our method achieves close performance to fully supervised counterpart with very few annotations. Across various segmentation heads, our method attains 98\% of the performance achievable under fully supervised conditions, utilizing just 2.2\% to 5\% annotations.
\begin{table}[t!]
\centering
\caption{Ablation Studies of Each Component.}
\label{tab:ablation_study}
\begin{adjustbox}{width=0.7\linewidth}
\begin{tabular}{cccc}
\toprule
\makecell[c]{Sampling} & \makecell[c]{DMC} & \makecell[c]{GTAV} & \makecell[c]{SYNTHIA}\\
\cmidrule(lr){1-2} \cmidrule(lr){3-3}  \cmidrule(lr){4-4} 
Baseline & \xmark & 66.2 & 68.2\\
\cmidrule(lr){1-2} \cmidrule(lr){3-3}  \cmidrule(lr){4-4} 
\multirow{2}{*}{K-Center Greedy} & \xmark & 69.4 & 70.7 \\
  & \cmark & 70.1 & 71.1 \\
\cmidrule(lr){1-2} \cmidrule(lr){3-3}  \cmidrule(lr){4-4} 
\multirow{2}{*}{Density-aware Greedy} & \xmark & 70.4 & 71.4 \\
  & \cmark & \textbf{71.1} & \textbf{72.1} \\
\bottomrule
\end{tabular}
\end{adjustbox}
\end{table}
\begin{table}[t!]
\centering
\caption{Bound and Core-set Loss comparison of K-Center Greedy and Density-aware Greedy.}
\label{tab:numerical_results}
\begin{adjustbox}{width=0.9\linewidth}
\begin{tabular}{cccc}
\toprule
\textbf{Algorithm} & $\delta$ & $max_{k\in \mathbf{s}} \delta_{\mathbf{s}} (k)$ & \textbf{Core-set Loss}\\
\midrule
K-Center Greedy & \textbf{0.132} & 0.364 & 0.646 \\
Density-aware Greedy & 0.176 & \textbf{0.124} & \textbf{0.550}\\
\bottomrule
\end{tabular}
\end{adjustbox}
\end{table}
\subsection{Ablation Studies}
\textbf{Effect of Dynamic Masked Convolution and Density-aware Greedy.}
we conduct ablation experiments on the GTAV $\rightarrow$ Cityscapes and SYNTHIA $\rightarrow$ Cityscapes tasks with a 2.2\% label budget to explore the impact of our proposed Dynamic Masked Convolution (DMC) and  Core-set selection, as shown in Table~\ref{tab:ablation_study}. The `Baseline` method uses only the entropy of the model output for active selection. The `K-Center Greedy' first filters candidate samples using entropy and then applies K-Center Greedy algorithm described in ~\cite{sener2017active} for active selection. When combined with DMC, the K-Center Greedy algorithm only utilizes DMC as a feature regularization technique. For our Density-aware Greedy method, removing DMC involves using the context model from ~\cite{minnen2018joint} in learned image compression to estimate density.

From the results in Table~\ref{tab:ablation_study}, the following observations can be made: (1) Our proposed Density-aware Greedy algorithm demonstrates robust performance improvements compared to directly applying the Core-set approach~\cite{sener2017active} in semantic segmentation. Compared to K-Center Greedy without DMC, our method achieves a mIOU improvement of 1.7 and 1.4 on GTAV and SYNTHIA, respectively. Even when DMC is replaced with a $5\times 5$ masked convolution, we still achieve mIOU improvements of 1.0 and 0.7 on GTAV and SYNTHIA, respectively. This highlights the significance of coverage density. (2) The proposed DMC is more suitable for estimating density compared to the context model originally used in learned image compression, resulting in an mIOU improvement of 0.7. (3) Introducing DMC solely as a feature regularization technique also improves model performance. For K-Center Greedy, it leads to mIOU improvement of 0.7 and 0.4, respectively.

\begin{table}[t!]
\centering
\caption{Comparison with Active Baselines.}
\label{tab:active_baselines}
\begin{adjustbox}{width=0.8\linewidth}
\begin{tabular}{lcc}
\toprule
\makecell[c]{Method} & \makecell[c]{GTAV} & \makecell[c]{SYNTHIA} \\
\cmidrule(lr){1-1} \cmidrule(lr){2-2} \cmidrule(lr){3-3}
RAND & 63.8 & 65.6\\
ReDAL \cite{wu2021redal} & 66.2 & 67.2 \\
BADGE \cite{ash2019deep} & 66.1 & 67.1 \\
ENT \cite{shen2017deep} & 66.2 & 68.2 \\
SCONF \cite{culotta2005reducing} & 66.5 & 68.4 \\
MARGIN \cite{wang2014new} & 66.1 & 68.0\\
Ours (w/o source data) & 69.1 & 70.0 \\
\textbf{Ours} & \textbf{71.1} & \textbf{72.1} \\
\bottomrule
\end{tabular}
\end{adjustbox}
\end{table}
Moreover, we conducted numerical experiments on the Cityscapes training set to validate that the proposed Density-aware Greedy algorithm reduces the new bound introduced in Eq. \ref{equ:tighter_upper_bound}. We compared the bounds $\delta$ from Eq. \ref{equ:coverage_area} and $max_{k\in \mathbf{s}} \delta_{\mathbf{s}} (k)$ from Eq. \ref{equ:tighter_upper_bound} for models trained with annotations selected using K-center Greedy and Density-aware Greedy. The results are shown in Table \ref{tab:numerical_results}. It is observed that the K-center Greedy algorithm results in a smaller $\delta$, yet a larger $max_{k\in \mathbf{s}} \delta_{\mathbf{s}} (k)$. Conversely, our proposed Density-aware Greedy algorithm results in a larger $\delta$, while yielding a smaller $max_{k\in \mathbf{s}} \delta_{\mathbf{s}} (k)$. Since $max_{k\in \mathbf{s}} \delta_{\mathbf{s}} (k)$ is a tighter bound, the actual Core-set Loss (cf. Eq. \ref{equ:core_set_target}) of the Density-aware Greedy algorithm is also smaller.

\textbf{Comparison with Common Active Learning Baselines.}
We also compared our method with other active learning methods, including random (RAND), uncertainty-based methods (ENT, SCONF and MARGIN), and hybrid methods (ReDAL and BADGE). The label budget is set to 2.2\%. It can be seen from Table~\ref{tab:active_baselines} that: (1) Active learning methods bring obvious performance gain over random selection. Compared with RAND, our method can achieve a significant mIOU improvement of 7.3 and 6.5. (2) Even without any source domain data, our method still outperforms commonly used active learning strategies by a margin over 2 mIOU, demonstrating the competitiveness of our method as an active learning approach.

\section{Conclusion} 
In this paper, we propose a Density-aware Core-set Selection method for active domain adaptive segmentation. We derive a tighter upper bound for the classical Core-set and identify that the model performance is closely related to the coverage sample distribution of selected samples. Further, we introduce a Proxy Density Estimator and develop a Density-aware Greedy algorithm to optimize the newly derived bound. Experiments demonstrate that the proposed method outperforms existing active learning and domain adaptation methods.

\section{Acknowledgements}
The paper is supported in part by the National Natural Science Foundation of China (No. 62325109, U21B2013, 61971277).

\bibliography{aaai24}

\appendix

\newtheorem{lemma}{Lemma}
\newpage
\section{Proof}
We first briefly review notation definitions. The data space is denoted as $\mathcal X$ and the label space is denoted as $\mathcal Y=\{ 
1, ..., C \}$, where $C$ is the number of classes. The training set $\{ x_t, y_t \}_{t\in [n]}$ is i.i.d. sampled from the space $\mathcal Z = \mathcal X \times \mathcal Y$. For the ease of representation, without the loss of generality, we represent the selected data points as $\tilde{\mathcal{X}}=\{\tilde{\mathbf{x}}_1,\tilde{\mathbf{x}}_2, \cdots, \tilde{\mathbf{x}}_b\}$ with the subscripts set denoted as $\mathbf{s}=\{1,2,\cdots, b\}$ and corresponding coverage area denoted as $\mathbb{N}_{\mathbf{s}}(k),k\in \mathbf{s}$. The class-wise label distribution is denoted as $\eta_c (\mathbf x)=p(y=c|\mathbf x)$ and the loss function is denoted as $l(\cdot, \cdot, A_{\mathbf s})$.
\subsection{Proof of Theorem 2}
Similar to the proof from Sener \& Savarese (2017), we first introduce the  Lemma from Berlind \& Urner (2015).
\begin{lemma}\label{lemma}
    Suppose $y$ conforms to Bernoulli distribution $\mathcal{B}(\tau)$ parameterized by $\tau$, then given $\tau,\tau' \in [0, 1]$, for any $y'\in\{0,1\}$,
    \begin{equation}
        p_{y\sim\mathcal{B}(\tau)}(y\neq y') \leq p_{y\sim\mathcal{B}(\tau')}(y\neq y') + |\tau - \tau'|.
    \end{equation}
\end{lemma}
Then we analyze the expected Core-set error of $\mathbb{E}_{y}\left[l(\mathbf{x},y;A_{\mathbf{s}})|\mathbf{x},\tilde{\mathcal{X}}\right]$ under the condition of selected set $\tilde{\mathcal{X}}$ and data point $\mathbf{x}$. We can expand the expectation in terms of condtioning on different classes:
\begin{small}
\begin{equation}\label{eq:expectation}
\mathbb{E}_{y}\left[l(\mathbf{x},y;A_{\mathbf{s}})|\mathbf{x},\tilde{\mathcal{X}}\right] = \sum_{c=1}^C {p_{y\sim\eta_c(\mathbf{x})}(y=1)l(\mathbf{x},c;A_{\mathbf{s}})}.
\end{equation}
\end{small}
Further, given $\tilde{\mathcal{X}}$ and $\mathbf{x}$, for a data point $\mathbf{x} \in \mathcal{X}$, we define its nearest point in selected $\tilde{\mathcal{X}}$ as:
\begin{equation}
    \pi(\mathbf{x}) = \arg\min_{\tilde{\mathbf{x}} \in \tilde{\mathcal{X}}}|\mathbf{x} - \tilde{\mathbf{x}}|,
\end{equation}
then with the help of Lemma~\ref{lemma}, the Eq~\ref{eq:expectation} can be relaxed as:
\begin{scriptsize}
\begin{align}
    & \mathbb{E}_{y}\left[l(\mathbf{x},y;A_{\mathbf{s}})|\mathbf{x},\tilde{\mathcal{X}}\right] = \sum_{c=1}^C {p_{y\sim\eta_c(\mathbf{x})}(y=1)l(\mathbf{x},c;A_{\mathbf{s}})} \notag \\
    & \leq \sum_{c=1}^C {\left[p_{y\sim\eta_c(\pi(\mathbf{x}))}(y=1)+|\eta_c(\mathbf{x})-\eta_c(\pi(\mathbf{x}))|\right]l(\mathbf{x},c;A_{\mathbf{s}})} \notag \\
    & = \sum_{c=1}^C {p_{y\sim\eta_c(\pi(\mathbf{x}))}(y=1)l(\mathbf{x},c;A_{\mathbf{s}})} + \sum_{c=1}^C |\eta_c(\mathbf{x})-\eta_c(\pi(\mathbf{x}))|l(\mathbf{x},c;A_{\mathbf{s}}) \notag \\
    & \leq \sum_{c=1}^C {p_{y\sim\eta_c(\pi(\mathbf{x}))}(y=1)l(\mathbf{x},c;A_{\mathbf{s}})} +\lambda^{\eta}LC|\mathbf{x}-\pi(\mathbf{x})| \label{ineq:lip1},
\end{align}
\end{scriptsize}
where the Eq~\ref{ineq:lip1} is obtained with the assumption that $\eta(\mathbf{x})$ is $\lambda^{\eta}$-Lipschitz and $|l(\cdot, \cdot; A_{\mathbf{s}})| \leq L$. On the other hand, with the assumption $l(\tilde{\mathbf{x}},y;A_{\mathbf{x}})=0 \quad \forall \tilde{\mathbf{x}} \in \tilde{\mathcal{X}}$ and $l(\cdot, \cdot; A_{\mathbf{s}})$ is $\lambda^l$-Lipschitz, we can obtain
\begin{small}
\begin{align}\label{ineq:lip2}
        l(\mathbf{x},c;A_{\mathbf{s}}) = l(\mathbf{x},c;A_{\mathbf{s}}) - l(\pi(\mathbf{x}),c;A_{\mathbf{s}}) \leq \lambda^l|\mathbf{x} - \pi(\mathbf{x})|.
\end{align}
\end{small}
By inserting Eq~\ref{ineq:lip2} into Eq~\ref{ineq:lip1}, we obtain the form of expected error as:
\begin{equation}
    \mathbb{E}_{y}\left[l(\mathbf{x},y;A_{\mathbf{s}})|\mathbf{x},\tilde{\mathcal{X}}\right] \leq (\lambda^l + \lambda^{\eta}LC)|\mathbf{x} - \pi(\mathbf{x})|.
\end{equation}
Next, we focus on a more general expectation error $\mathbb{E}_{\mathbf{x}, y}\left[l(\mathbf{x},y;A_{\mathbf{s}})|\tilde{\mathcal{X}}\right]$ conditioned on the selected samples $\tilde{\mathcal{X}}$, since points in $\tilde{\mathcal{X}}$ are drawn independently, thus $p(\mathbf{x}|\tilde{\mathcal{X}})=p(\mathbf{x})$, and the expectation can be expressed as:
\begin{scriptsize}
\begin{align}
    & \mathbb{E}_{\mathbf{x}, y}\left[l(\mathbf{x},y;A_{\mathbf{s}})|\tilde{\mathcal{X}}\right] = \int_{\mathbf{x}} p(\mathbf{x})\mathbb{E}_{y}\left[l(\mathbf{x},y;A_{\mathbf{s}})|\mathbf{x},\tilde{\mathcal{X}}\right] d\mathbf{x} \notag \\
    & \leq \left(\lambda^l + \lambda^{\eta}LC\right) \int_{\mathbf{x}}|\mathbf{x} - \pi(\mathbf{x})|p(\mathbf{x})d\mathbf{x} \notag \\
    & = \left(\lambda^l + \lambda^{\eta}LC\right) \int_{\mathbf{x}}|\mathbf{x} - \pi(\mathbf{x})|\sum_{k\in\mathbf{s}}p(\mathbf{x}|\pi(\mathbf{x})=\tilde{\mathbf{x}}_k)p(\pi(\mathbf{x})=\tilde{\mathbf{x}}_k)d\mathbf{x} \notag \\
    & = \left(\lambda^l + \lambda^{\eta}LC\right) \sum_{k\in\mathbf{s}}{p(\pi(\mathbf{x})=\tilde{\mathbf{x}}_k)}\int_{\mathbf{x}}|\mathbf{x} - \tilde{\mathbf{x}}_k|p(\mathbf{x}|\pi(\mathbf{x})=\tilde{\mathbf{x}}_k)d\mathbf{x} \notag \\
    & = \left(\lambda^l + \lambda^{\eta}LC\right) \sum_{k\in\mathbf{s}}{p(\pi(\mathbf{x})=\tilde{\mathbf{x}}_k)}\mathbb{E}_{\mathbf{x}\sim \mathbb{N}_{\mathbf{s}}(k)}[|\mathbf{x}-\tilde{\mathbf{x}}_k|] \notag \\
    & \leq \left(\lambda^l + \lambda^{\eta}LC\right) \max_{k}\mathbb{E}_{\mathbf{x}\sim \mathbb{N}_{\mathbf{s}}(k)}[|\mathbf{x}-\tilde{\mathbf{x}}_k|].
\end{align}
\end{scriptsize}
Next, in terms of  Hoeffding’s inequality, since $l(\mathbf{x},y;A_{\mathbf{s}})\leq L$, given a small error value $\epsilon$, and $n$ samples $\mathbf{x}_t$ from $\mathcal{X}$, we have:
\begin{scriptsize}
\begin{equation}\label{eq:hoffding}
    p\left(\frac{1}{n}\sum_{t}{l(\mathbf{x}_t, y; A_{\mathbf{s}})} - \mathbb{E}_{\mathbf{x}, y}\left[l(\mathbf{x},y_t;A_{\mathbf{s}})|\tilde{\mathcal{X}}\right]\leq\epsilon\right) \geq 1 - \exp\left(-\frac{2n\epsilon^2}{L^2}\right).
\end{equation}
\end{scriptsize}
By representing the lower bound probability in Eq~\ref{eq:hoffding} as $1-\gamma$, we can inversely solve the error value,
\begin{align}
    \gamma =\exp\left(-\frac{2n\epsilon^2}{L}\right) \Longrightarrow \epsilon = \sqrt{\frac{L^2\ln{(1/\gamma)}}{2n}}.
\end{align}
Considering $l(\tilde{\mathbf{x}},y;A_{\mathbf{s}})=0 \quad \forall \tilde{\mathbf{x}} \in \tilde{\mathcal{X}}$, we represent $\delta_{\mathbf{s}}(k)=\mathbb{E}_{\mathbf{x}\sim\mathbb{N}_{\mathbf{s}}(k)}[|\mathbf{x}-\mathbf{\tilde{x}}_k|]$ and can conclude that with probability at least $1-\gamma$,
\begin{small}
\begin{align}
    & |\frac{1}{n}\sum_{t}{l(\mathbf{x}_t, y; A_{\mathbf{s}})} - \frac{1}{|\mathbf{s}|}\sum_{k}{l(\tilde{\mathbf{x}}_k, y; A_{\mathbf{s}})}| \notag \\
    & \leq \mathbb{E}_{\mathbf{x}, y} \left[l(\mathbf{x},y;A_{\mathbf{s}})|\tilde{\mathcal{X}}\right] + \epsilon \notag \\
    & \leq \max_{k\in \mathbf{s}}\delta_{\mathbf{s}}(k) \cdot (\lambda^l+\lambda^{\eta}LC) + \sqrt{\frac{L^2 \ln{(1/ \gamma)}}{2n}}
\end{align}
\end{small}

\subsection{Proof of Theorem 3}

To prove Theorem 3, we are essentially proving the following inequality
\begin{equation}
    \max_{k\in \mathbf{s}}\mathbb{E}_{\mathbf{x}\sim\mathbb{N}_{\mathbf{s}}(k)}[|\mathbf{x}-\tilde{\mathbf{x}}_k|] \leq \delta = \max_{k\in \mathbf{s}}\max_{\mathbf{x}\in\mathbb{N}_{\mathbf{s}}(k)}|\mathbf{x}-\tilde{\mathbf{x}}_k|.
\end{equation}
For the ease of representation, we define the optimal index as
\begin{align}
    k_1 &=\arg\max_{k\in \mathbf{s}}\mathbb{E}_{\mathbf{x}\sim\mathbb{N}_{\mathbf{s}}(k)}[|\mathbf{x}-\tilde{\mathbf{x}}_k|] \\
    k_2 &=\arg\max_{k\in \mathbf{s}}\max_{\mathbf{x}\in\mathbb{N}_{\mathbf{s}}(k)}|\mathbf{x}-\tilde{\mathbf{x}}_k|.
\end{align}
This inequality can be verified under different cases

\noindent\textbf{Case1.} When $k_1 = k_2 = k^*$, it is easy to verify that
\begin{equation}
\mathbb{E}_{\mathbf{x}\sim\mathbb{N}_{\mathbf{s}}(k^*)}[|\mathbf{x}-\tilde{\mathbf{x}}_{k^*}|] \leq \max_{\mathbf{x}\in\mathbb{N}_{\mathbf{s}}(k^*)}|\mathbf{x}-\tilde{\mathbf{x}}_{k^*}|.
\end{equation}

\noindent\textbf{Case2.} When $k_1 \neq k_2$, we have
\begin{align}
    & \mathbb{E}_{\mathbf{x}\sim\mathbb{N}_{\mathbf{s}}(k_1)}[|\mathbf{x}-\tilde{\mathbf{x}}_{k_1}|] = \int_{\mathbf{x}}{|\mathbf{x}-\tilde{\mathbf{x}}_{k_1}|p(\mathbf{x}|\pi(\mathbf{x})=\tilde{\mathbf{x}}_{k_1})}d\mathbf{x} \notag \\ 
    & \leq \max_{\mathbf{x}\in\mathbb{N}_{\mathbf{s}}(k_1)}[|\mathbf{x}-\tilde{\mathbf{x}}_{k_1}|] \leq \max_{\mathbf{x}\in\mathbb{N}_{\mathbf{s}}(k_2)}[|\mathbf{x}-\tilde{\mathbf{x}}_{k_2}|].
\end{align}
Hence the inequality is proved.

\begin{figure}[t!]
    \centering
    \includegraphics[width=\linewidth]{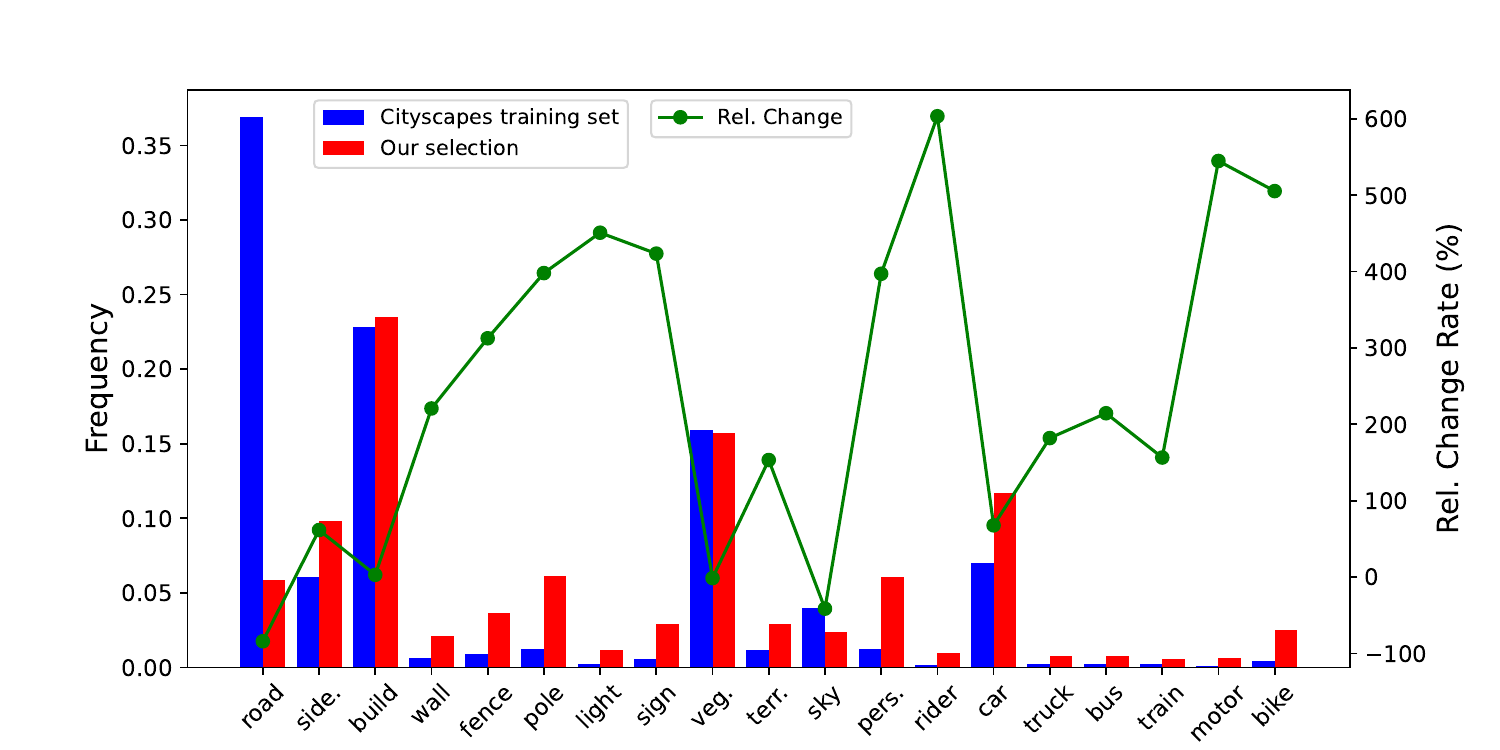}
    \caption{The class distribution of the selected regions on the cityscapes training dataset.}
    \label{fig:label_distribution}
\end{figure}

\section{Label Distribution}
Although we have not explicitly designed a mechanism for class balancing, our Adaptive Radius Core-set algorithm tends to select diverse data, which implicitly preserves the balance among different semantic categories to some extent, as shown in Fig.~\ref{fig:label_distribution}. We observe that our method selects a smaller proportion of samples for the dominant classes in the training set, while it selects a higher proportion for the minority classes. This also conforms to our experimental results that our method performs well on the minority classes.

\section{Impact of Candidate Filtering Method}
\begin{figure}[t!]
    \centering
    \includegraphics[width=0.6\linewidth]{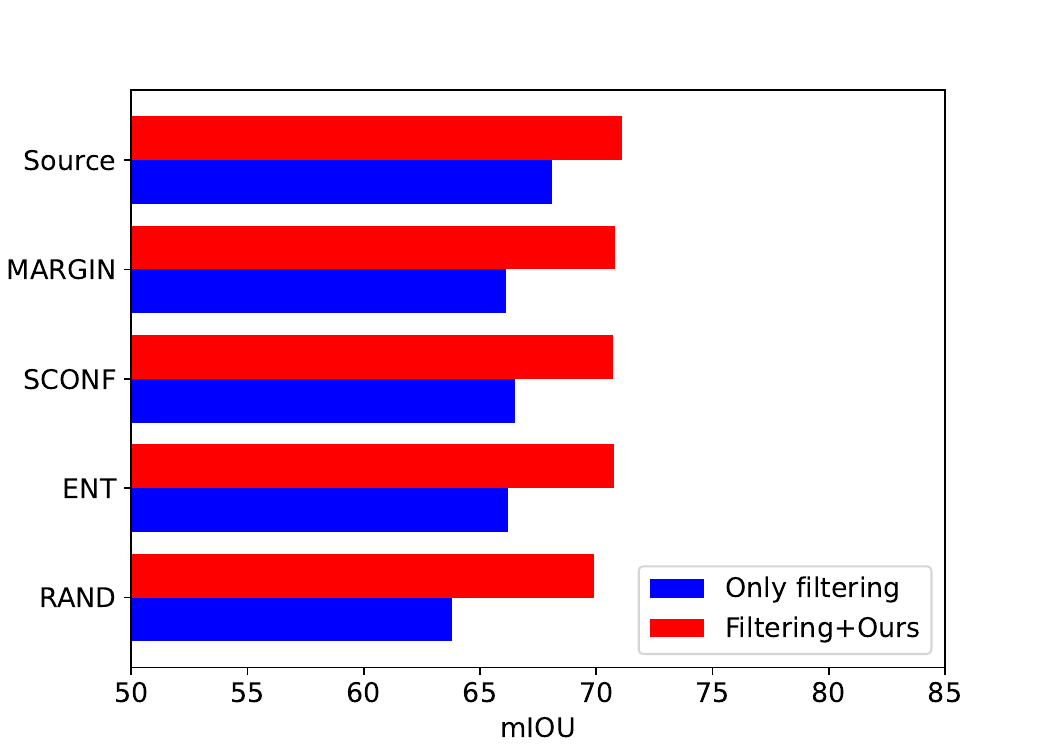}
    \caption{Impact of using different filtering methods for candidate selection.}
    \label{fig:filter_comparison}
\end{figure}

In addition, we study the impact of using different metrics for candidate selection, which is illustrated in Fig.~\ref{fig:filter_comparison}. It can be observed that with common uncertainty metrics (e.g. ENT, SCONF, MARGIN) as filtering method, our method achieves robust performance improvement of over 4 mIOU. Moreover, source-aided candidate filtering achieves best performance among all filtering methods, demonstrating the validity of source data for pre-filtering candidate samples.

\begin{figure*}[t!]
\centering
\includegraphics[width=\linewidth]{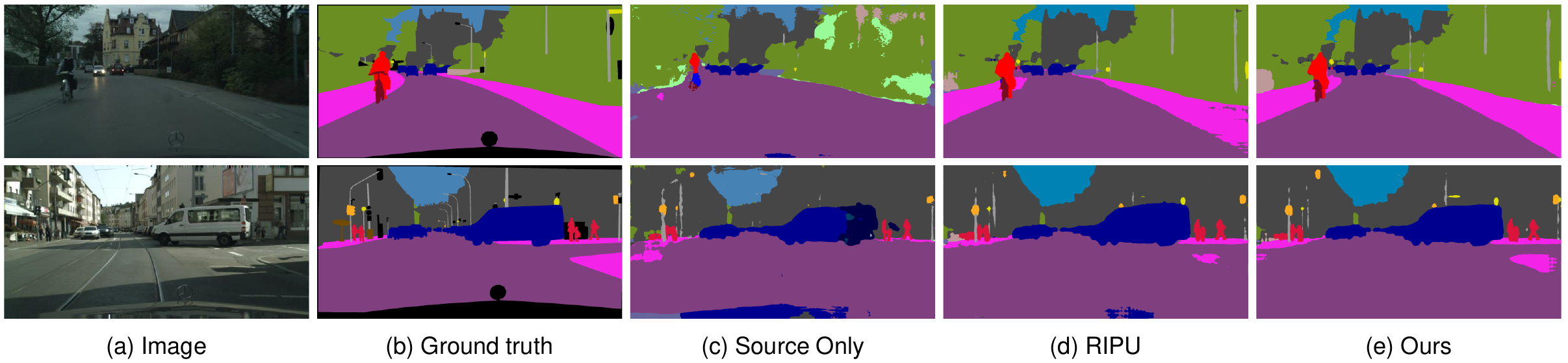}
\caption{Visualization of segmentation results on GTAV $\rightarrow$ Cityscapes.}
\label{fig:segmentation_results}
\end{figure*}
\begin{figure*}[t!]
\centering
\includegraphics[width=\linewidth]{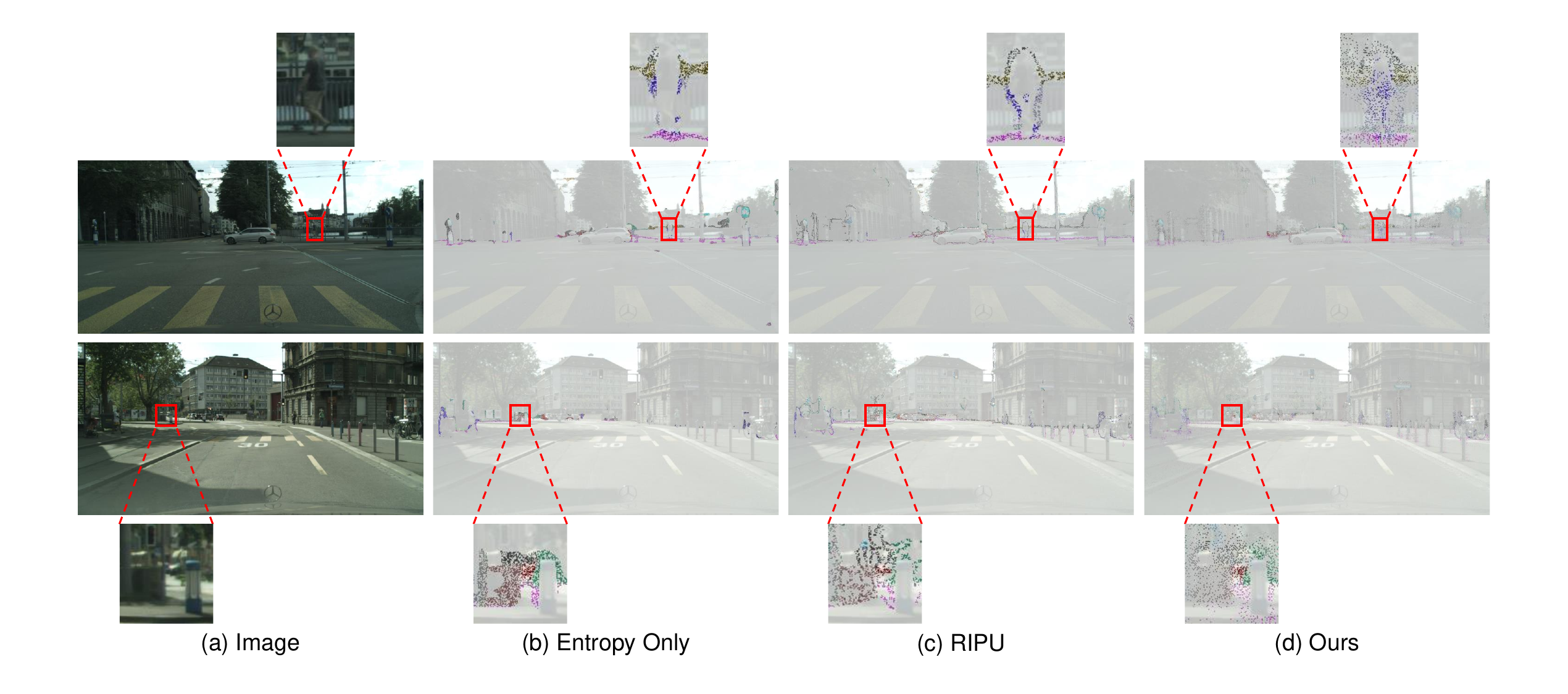}
\caption{Visualization of queried regions on GTAV $\rightarrow$ Cityscapes. The label budget is 2.2\%.}
\label{fig:queried_regions}
\end{figure*}
\section{Results with Different Density Estimator}
We conducted additional experiments on the GTAV $\rightarrow$ Cityscapes and SYNTHIA $\rightarrow$ Cityscapes tasks with a 2.2\% label budget to investigate the influence of different density estimators on our method. The results are presented in Tab. ~\ref{tab:density_estimator}. In Tab. ~\ref{tab:density_estimator}, `$3\times 3$ masked conv' and `$5\times 5$ masked conv' refer to using convolutions with the kernel center masked out as density estimators, similar to the context model in learned image compression. We also experimented with different kernel sizes for the dynamic convolution in DMC, resulting in the `3x3 DMC' and `5x5 DMC' settings.

From the results in Table~\ref{tab:density_estimator}, we can observe that: (1) Our proposed Density-aware Greedy algorithm exhibits robustness to the structure of the density estimator. The variation in performance resulting from the density estimator's structure is within 0.7 mIOU. (2) The proposed DMC is well-suited for density estimation, and a relatively larger kernel size enhances the accuracy of density estimation. The performance of `$5\times 5$ DMC' shows an mIOU improvement of 0.6 on GTAV $\rightarrow$ Cityscapes and 0.3 on SYNTHIA $\rightarrow$ Cityscapes compared to `$3\times 3$ DMC'.

\begin{table}[h!]
\centering
\caption{Comparison of different density estimators.}
\begin{tabular}{ccc}
\toprule
Density Estimator & GTAV & SYNTHIA\\
\hline
$3\times 3$ masked conv & 70.3 & 71.2 \\
$5\times 5$ masked conv & 70.4 & 71.4 \\
$3\times 3$ DMC & 70.4 & 71.5 \\
$5\times 5$ DMC & \textbf{71.1} & \textbf{72.1} \\
\bottomrule
\end{tabular}
\label{tab:density_estimator}
\end{table}

\begin{table}[h]
\centering
\caption{Source-free Comparision.}
\label{tab:source_free}
\begin{adjustbox}{width=\linewidth}
\begin{tabular}{l|ccc}
    \toprule
    \makecell[c]{Method} & Budget & GTAV & SYNTHIA \\
    \cmidrule(lr){1-1} \cmidrule(lr){2-2}  \cmidrule(lr){3-3} \cmidrule(lr){4-4} 
    URMA \cite{wang2021uncertainty} & - & 45.1 & 39.6 \\
    LD \cite{you2021domain} & - & 45.5 & 42.6 \\
    SFDA \cite{kundu2021generalize} & - & 53.4 & 52.0 \\
    RIPU \cite{xie2022towards} & 2.2\% & 67.1 & 68.7 \\
    \textbf{Ours} & 2.2\% & \textbf{70.3} & \textbf{71.5} \\
    \bottomrule
\end{tabular}
\end{adjustbox}
\end{table}
\section{Extension to Source-free Scenario.}
We verify the effectiveness of our method in the source-free scenario where only a pre-trained model from the source domain is available, and the source domain's data and labels cannot be obtained. Our method outperforms the previous state-of-the-art RIPU by a large margin in Table~\ref{tab:source_free}, demonstrating the superiority of our method in source-free domain adaptation.

\section{Visualization of Feature Distribution}
\begin{figure}[h!]
\centering
\includegraphics[width=0.8\linewidth]{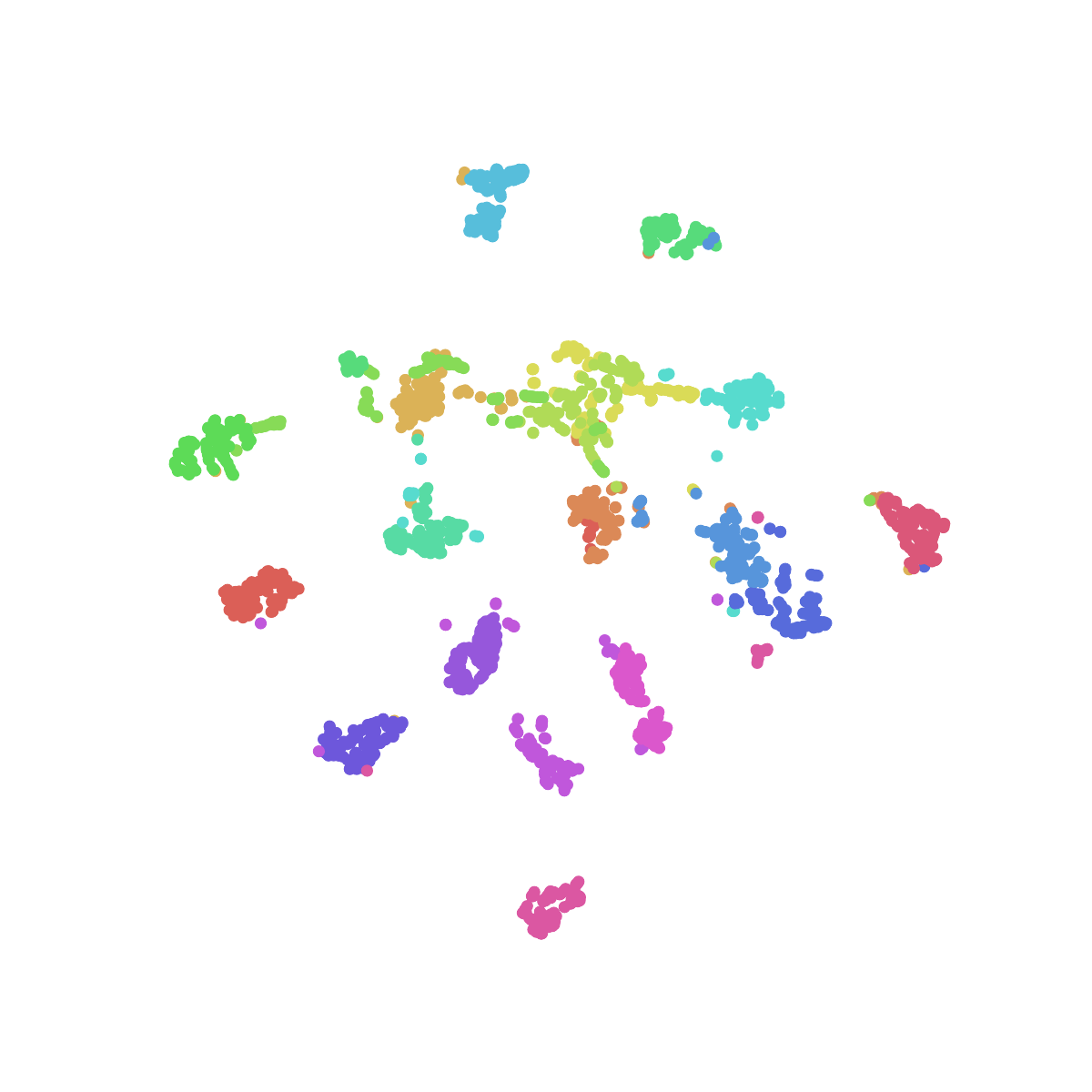}
\caption{T-SNE visualization of feature distribution.}
\label{fig:tsne}
\end{figure}
We visualize the distribution of features used for selection by the Density-aware Greedy algorithm in Fig.~\ref{fig:tsne}. T-SNE is adopted as the dimension reduction method. Two observations can be drawn: (1) The distribution of features in the feature space is not uniform, but there are dense and sparse regions. Therefore, the original Core-set method based on the K-center Greedy algorithm cannot optimize the new upper bound we derived well. This makes it necessary to propose the Density-aware Greedy algorithm. (2) Our proposed method can distinguish the features of different categories well, so as to achieve the target of pixel-level classification.

\section{Qualitative Results}
We present visualizations of the regions queried for labeling by different active learning methods in Fig.~\ref{fig:queried_regions}. It is evident that the `Uncertainty Only' method and RIPU tend to focus their queries on the image edges. In contrast, our method exhibits more diverse queried regions. While regions with high uncertainty near the edges are prioritized, a small number of labels are also queried in non-edge regions.

To provide further insights, we visualize the segmentation results of various methods in Fig.~\ref{fig:segmentation_results}. It can be seen that our method performs better in segmenting minority classes.

\section{Hyperparameter Sensitivity}
The hyperparameters we tune mainly include $\beta$ and $\tau$ within $\mathbf{D}_{i,j} = \beta\exp\left(-{||\hat{\mathbf{F}}_{:,i,j} - \mathbf{F}_{:,i,j}||_2^2/\tau}\right)$. The purpose of introducing $\beta$ is to prevent the elimination of small differences in density due to floating-point precision. Since it linearly scales the density, the algorithm is not sensitive to it. For another thing, a smaller $\tau$ leads to more significant differences in density corresponding to the same reconstruction errors. In our experiments, we found that values of $\tau$ between 0.15 and 0.3 all lead to a reduction in the bound and an improvement in mIOU. Thus the algorithm is not very sensitive to $\tau$ either.

\section{Combination with UDA Methods}
Our approach can complement exising UDA methods effectively. Take MIC~\cite{hoyer2023mic} as an example, by allowing MIC use active-selected labels, our method achieves an mIOU of 77.6. Our method brings an improvement of 3.9 mIOU compared with random selection on the GTAV$\rightarrow$Cityscapes task.

\section{Assessing the Quality of the Density Estimator}
We can access the quality of the trained density estimator through a calibration experiment. We first sample samples with densities approximately following $U(0.5, 3)$. Then we applied linear regression to average radial distance and the inverse of density, resulting in an $R^2$ value of 0.72 for our DMC. This experiment demonstrate that our density estimator effectively captures the relationship between density and distance.

\section{More Implementation Details}
In this paper, we conducted all experiments using a Tesla V100 GPU with pytorch 1.6.0. Before feeding images into the network, the source image is resized to $1280\times 720$, and the target image is resized to $1280\times 640$. Our training settings are aligned with RIPU, i.e. using the SGD optimizer with an initial learning rate of $2.5\times 10^{-4}$, a batch size of 2 and adopt prediction consistency loss and negative learning loss for fair comparison. We train for 40K iterations in all experiments. The complete training and active selection process took approximately two days. After 18000 iterations, we prevent the gradients of the proxy density estimator from being propagated back to the backbone, ensuring that it does not affect the convergence of the model. Additionally, since we estimate candidate density in the feature space, we employed bilinear interpolation to upsample the feature $\mathbf{f}$ and density $\mathbf{d}$ of candidate samples to match the size of the original image before active selection.

\end{document}